\documentclass[10pt,twocolumn,letterpaper]{article}

\usepackage{panda}              
\usepackage{amssymb} 
\usepackage{multirow}
\usepackage{pifont}
\usepackage{algorithm}
\usepackage{algpseudocode}
\usepackage{enumitem}
\usepackage[dvipsnames]{xcolor}

\definecolor{pandablue}{rgb}{0.21,0.49,0.74}
\usepackage[pagebackref,breaklinks,colorlinks,citecolor=pandablue]{hyperref}

\title{DARNet: Bridging Domain Gaps in Cross-Domain Few-Shot Segmentation with Dynamic Adaptation}

\author{\small Haoran Fan\\
\small UNSW\\
{\tt\footnotesize fanhaoran68@gmail.com}
\and
\small Qi Fan\\
\small HKUST\\
{\tt\footnotesize qfanaa@cse.ust.hk}
\and
\small Maurice Pagnucco\\
\small UNSW\\
{\tt\footnotesize morri@cse.unsw.edu.au}
\and
\small Yang Song\\
\small UNSW\\
{\tt\footnotesize yang.song1@unsw.edu.au}
}

\begin{document}
\maketitle
\begin{abstract}
Few-shot segmentation (FSS) aims to segment novel classes in a query image by using only a small number of supporting images from base classes. However, in cross-domain few-shot segmentation (CD-FSS), leveraging features from label-rich domains for resource-constrained domains poses challenges due to domain discrepancies. This work presents a Dynamically Adaptive Refine (DARNet) method that aims to balance generalization and specificity for CD-FSS. Our method includes the Channel Statistics Disruption (CSD) strategy, which perturbs feature channel statistics in the source domain, bolstering generalization to unknown target domains. Moreover, recognizing the variability across target domains, an Adaptive Refine Self-Matching (ARSM) method is also proposed to adjust the matching threshold and dynamically refine the prediction result with the self-matching method, enhancing accuracy. We also present a Test-Time Adaptation (TTA) method to refine the model's adaptability to diverse feature distributions. Our approach demonstrates superior performance against state-of-the-art methods in CD-FSS tasks. 
\end{abstract}    
\section{Introduction}
\label{sec:intro}
Semantic segmentation tasks aim to divide images into distinct segments, with each segment representing a clear and meaningful region. Many deep learning-based semantic segmentation methods \cite{chen2017deeplab,long2015fully,zhao2017pyramid} have been developed, setting new standards in the field. However, a big challenge remains: these models require a lot of detailed, pixel-level annotated training data. Collecting such detailed labeled datasets is arduous and time-intensive. This challenge thus led to the development of the few-shot segmentation approach. 
Few-shot segmentation (FSS) aims to segment novel classes in a query image by using only a small number of supporting images with annotations.
The challenge arises from the limited set of supported images for each class, typically preset to {1, 3, 5, 10}, which contrasts with the huge number of query images. Therefore, it becomes very challenging to accurately represent the nuances of target classes in query images with limited few-shot support. To address this issue, various methods have been introduced, including altering the matching mechanism \cite{zhuge2021deep, yang2020brinet, liu2020dynamic} or crafting \cite{li2021adaptive, ouyang2020self, yang2021mining} representative prototypes for the query dataset. However, these strategies often fall short of bridging the appearance discrepancies between the support and query images. To address this issue, SSPNet \cite{fan2022self} presents an approach that leverages query features for support, which more effectively mitigates this challenge. 

\begin{figure}[t]
  \centering
  \includegraphics[width=1.0\linewidth]{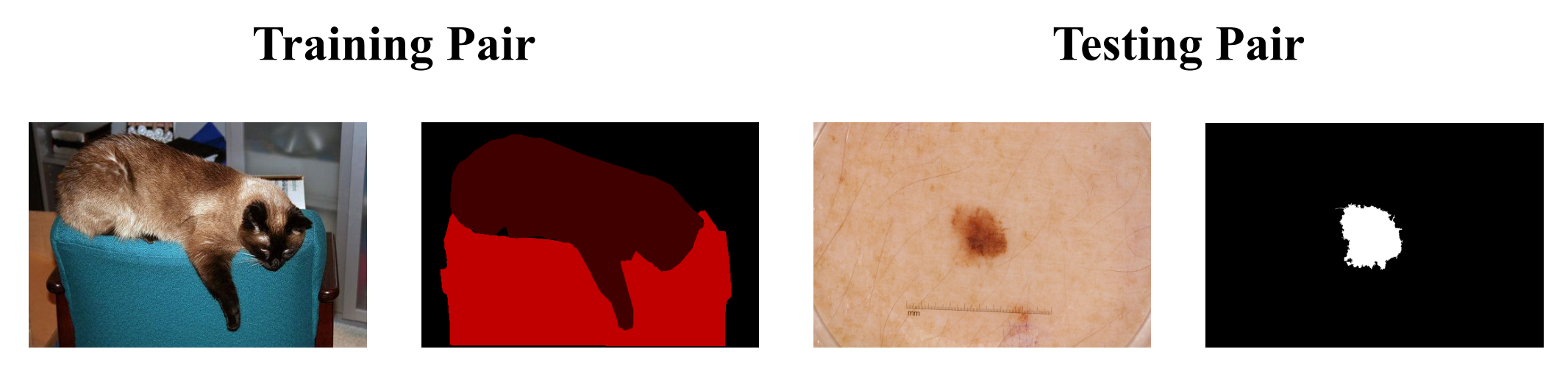}
  \caption{
   The training set and testing set in the CD-FSS task come from different domains, so their feature distributions and label spaces (classes) are different.}
  \label{fig:cdfsstask}
\end{figure}

Training models for FSS tasks typically require datasets from the same domain, this is often not viable for applications with scarce annotations. A new method called cross-domain few-shot segmentation (CD-FSS) has thus been proposed. As shown in Fig.~\ref{fig:cdfsstask}, most strategies \cite{lei2022cross,  lu2022cross, huang2023restnet} for CD-FSS tasks leverage meta-knowledge from label-rich source domains, such as PASCAL VOC \cite{everingham2010pascal}, and then apply it to resource-constrained target domains of entirely different images and object categories, such as images showing different classes of skin lesions. Therefore, the primary challenge lies in effectively transferring features learned from the source domain to the very different target domain. However, since the query image is not accessible during the fine-tuning phase of testing, and these methods do not account for the intra-class inconsistency between the support and query images, current CD-FSS methods show limited generalizability on the target domain.

It has been shown that using query features to self-match images reduces the intra-class inconsistency between support and query images \cite{fan2022self}. However, in the CD-FSS task, a significant domain gap might exist between source and target data. Relying solely on self-matching for cross-domain tasks can lead to overfitting in the source domain. To improve the model's generalization from the source domain, in this work, we introduce a Channel Statistics Disruption (CSD) strategy allowing for a better adaptation from the source domain to various target domains.
On the other hand, stronger generalization can compromise fitting within specific domains. Therefore, to improve the model's fit across multiple target domains, we employ a Test-Time Adaptation (TTA) method and introduce a Distribution Alignment Module (DAM). This module prioritizes features relevant to the target task and de-emphasizes less critical ones. During testing, while the parameters of the backbone remain fixed, we fine-tune the DAM parameters to accentuate inter-class relationships of crucial features within the target domain, thus enhancing the model's adaptability and recognition of the target domain's feature distribution.

When performing self-matching on target domain data, we recognize that both the similarities between foreground and background of images and the disparity between support and query images can vary across different target domains. Consequently, relying on a singular, manually set matching threshold is unlikely to be universally effective for all target domain datasets. We propose an Adaptive Refine Self-Matching (ARSM) method to dynamically adjust the matching threshold and refine prediction results with the self-matching method in a closed loop according to each episode data of the target domain during testing.  We also train the learnable parameters of ARSM based on each target domain episode when performing TTA. For experiments, we train our model on the PASCAL VOC 2012 \cite{everingham2010pascal} and SBD \cite{hariharan2011semantic} datasets and test it on the DeepGlobe \cite{demir2018deepglobe}, ISIC \cite{codella2019skin, tschandl2018ham10000}, Chest X-Ray \cite{candemir2013lung, jaeger2013automatic}, and FSS-1000 datasets \cite{li2020fss}. Our contributions are summarized as follows:

\begin{itemize}[align=left, left=0pt, label=--, itemindent=0pt, labelsep=1em]

    \item  We propose our Dynamically Adaptive Refine (DARNet) method specifically curated to bolster the model's fitting capability across multiple target domains, striking a balance between generalization and specificity, and its performance on multiple CD-FSS tasks is higher than the current state-of-the-art methods.

    \item We introduce the CSD strategy to enhance generalizability by effectively mapping features from the source domain to an unknown target domain.

    \item  A Test-Time Adaptation method specifically adapted to CD-FSS task is proposed to improve the model's adaptability of diverse feature distributions.

    \item  An ARSM method is designed to dynamically adjust the matching threshold and refine prediction with the self-matching method results in a closed loop, optimizing accuracy across diverse target domain datasets.

\end{itemize}

\section{Related Work}

\textbf{Cross-domain semantic segmentation.} Recent work on cross-domain semantic segmentation are primarily bifurcated into two major research trajectories. The first is \textit{domain adaptation} (DA). These methods have access to target domain data during training and use these data to promote the model's adaptation to the target domain distribution and mitigate the effects of domain shift. The most common method is \textit{adversarial training} \cite{chen2017no, chen2018road, hung2018adversarial, tsai2019domain}, which uses generative adversarial networks (GANs) or other adversarial loss functions to make the distributions of the source and target domains closer. 
The second is \textit{domain generalization} (DG). Different from the DA method, the DG method does not use the data of the target domain. Recent DG methods can be mainly divided into two types: normalization and randomization according to the transformation method of source domain features. Normalization methods \cite{pan2018two, choi2021robustnet, xu2022dirl, peng2022semantic} focus on eliminating domain-specific styles to capture features that are invariant across different styles. Randomization-based methods, on the other hand, aim to prepare the model for uncertainty in the target domain by generating a sequence of image styles from the source domain. For example, some methods of data augmentation \cite{peng2021global, zhong2022adversarial}  and some methods of manipulating data in feature space \cite{zhao2022style}.
For the CD-FSS task, the target domain is inaccessible when training the backbone, and only a few support images in the target domain are available for adapting the segmentation model to the novel classes.

\textbf{Few-shot Semantic Segmentation.} Most few-shot segmentation methods are designed based on the idea of metric meta-learning. These methods can be divided into metric-based and relation-based methods according to whether their measurement tools have parameters that need to be learned. The relation-based few-shot segmentation methods (e.g. CaNet \cite{zhang2019canet}, RPMM \cite{yang2020prototype}, PGNet \cite{zhang2019pyramid}, PFENet \cite{tian2020prior} and HSNet \cite{min2021hypercorrelation}) based on the parameter structure are similar to the traditional semantic segmentation method, which is composed of encoder and a feature decoder. This method usually freezes the weights of the backbone network and trains the feature comparison module and feature decoder. On the other hand, the metric-based few-shot segmentation methods (e.g. PANet \cite{wang2019panet}, AMP \cite{siam2019amp}) use the idea of the prototype network to extract the prototype of the category first, and then use non-parametric measurement tools such as measurement functions to measure and classify. The RePRI \cite{boudiaf2021few} method uses a unique transduction inference method, which can complete the inference in the test phase without using meta-learning. 
However, these methods do not consider the inter-domain gap problem, which will lead to poor model performance during the inference stage. Our method solves the inter-domain gap problem of CD-FSS by generalizing the features of the source domain and adapting them to the target domain. 

\textbf{Cross-domain Few-shot Semantic Segmentation.} The difference between the CD-FSS task and the ordinary FSS task is that the target domain data set of the CD-FSS task and the source domain data set not only have different classes but also completely different feature distributions. To address the feature disparities, RTD \cite{wang2022remember} builds a meta-memory to collect domain-specific information between source domain instances and continuously registers styled domain information into the meta-memory during the training process. Then, the stored memory is loaded into the source and target domains to enhance the model's generalization ability. PATNet \cite{lei2022cross} designs a pyramid module to convert the features learned by the model from the source domain into domain-agnostic features. 
However, these methods do not explicitly address the intra-class disparity between the support and query images, so the inference effect is reduced during testing. Our method solves this problem through adaptive self-matching.


\begin{figure*}[t]
  \centering
  \includegraphics[width=1.0\linewidth]{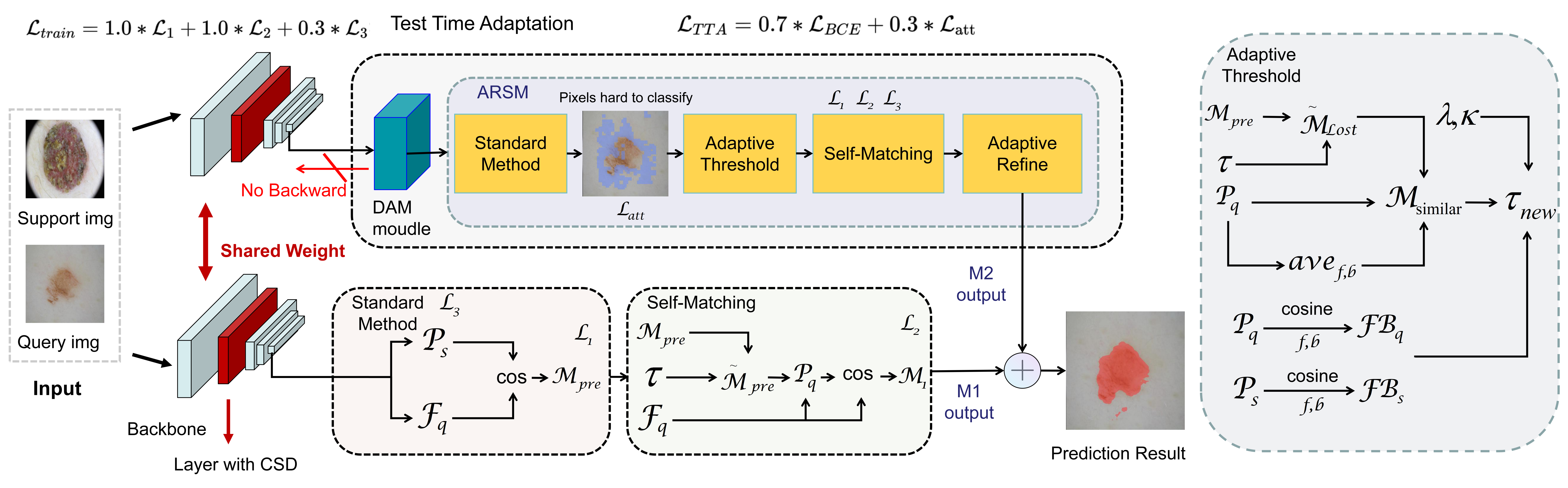}
  \caption{
   Overview of our method in a 1-way 1-shot example. After obtaining the features of the input image, we use the CSD strategy in the backbone to generalize the features and introduce the ARSM strategy to match the prediction results in the second branch. A TTA approach is also developed to finetune the parameters in the DAM module and the ARSM strategy during the testing phase. }
  \label{fig:image1}
\end{figure*}
\section{Methods}
In CD-FSS tasks, as Fig.~\ref{fig:cdfsstask} shows, we consider a source domain $(\mathcal{X}_s, \mathcal{Y}_s)$ and a target domain $(\mathcal{X}_t, \mathcal{Y}_t)$ with non-overlapping input distributions ($\mathcal{X}_{.}$) and label spaces ($\mathcal{Y}_{.}$), formalized as $\mathcal{X}_s \neq \mathcal{X}_t$ and $\mathcal{Y}_s \cap \mathcal{Y}_t = \emptyset$.
CD-FSS aims to train a model on the source domain $(\mathcal{X}_s, \mathcal{Y}_s)$ to segment new classes in the target domain $(\mathcal{X}_t, \mathcal{Y}_t)$ with only a few annotated samples per class. Training set $\mathcal{D}_{\text{train}}$ and testing set $\mathcal{D}_{\text{test}}$ are formed from $(\mathcal{X}_s, \mathcal{Y}_s)$ and $(\mathcal{X}_t, \mathcal{Y}_t)$, respectively, with episodes of $N$-way $K$-shot task. An episode comprises a support set $\mathcal{S} = \{(\mathcal{I}_s^i, \mathcal{M}_s^i)\}_{i=1}^{N \times K}$ and a query set $\mathcal{Q} = \{(\mathcal{I}_q^i, \mathcal{M}_q^i)\}_{i=1}^{Q}$, where $\mathcal{I}$ denotes an image, $\mathcal{M}$ its mask, and $Q$ the query samples count.  The support and query sets of each episode for both $\mathcal{D}_{\text{train}}$ and $\mathcal{D}_{\text{test}}$ are constructed as follows. From the corresponding dataset, we randomly select $N$ classes, $K$ examples for each class to form the support set,  another $Q$ different examples to form the query set, ensuring there is no overlap between support and query samples within the same episode \cite{vinyals2016matching}. The model is trained on $\mathcal{D}_{\text{train}}$ without exposure to target data $\mathcal{D}_{\text{test}}$. During testing, the trained model is presented with query and support sets from $\mathcal{D}_{\text{test}}$ for model evaluation. Following existing studies \cite{lei2022cross}, we performed 1-way 1-shot and 1-way 5-shot segmentations in this study.  

In standard FSS approaches, support and query images $\{I_s, I_q\}$ pass through a shared-weight backbone to extract features $\{F_s, F_q\} \in \mathbb{R}^{C \times H \times W}$. Using ground truth mask $M_s$, mask average pooling is applied on $F_s$ to generate foreground and background support prototypes $P_s = \{ P_{s,f}, P_{s,b} \} \in \mathbb{R}^{C \times 1 \times 1}$. Foreground and background distance maps $D = \{ D_f, D_b \}$ are then formed based on cosine similarity between $P_s$ and $F_q$, and the final prediction $M_{0}$ is obtained by computing softmax on $D$. However, this method doesn't address the intra-class inconsistency between support and query images in FSS tasks and the cross-domain gap in CD-FSS tasks.

As shown in Fig.~\ref{fig:image1}, in order to solve these two problems, our DARNet proposes a dual-branch solution. In the first branch, we use a self-matching method in both training and testing phases to generate query prototype $P_{q}^{*}$ by combining the prediction masks $M_{0}$ and $F_{q}$ generated by the method similar to the standard FSS method.  Then we use $P_{q}^{*}$ to self-match the query feature $F_{q}$ again to obtain the prediction result $M_{1}$. In the second branch, we place a Domain Adaptation Module (DAM) after the backbone and then apply the ARSM strategy. ARSM adaptively adjusts the threshold for self-matching to continuously refine the prediction results. DAM is an attention-based module to highlight  important features, which is fine-tuned during test-time adaptation to further address the problem of cross-domain gaps. The prediction result from the second branch is denoted as $M_2$.  Also, during the training phase, we apply the CSD strategy in the shared backbone to enhance the model’s generalization ability by perturbing the channel statistics in each block of the low layer. Finally, we obtain the final result by a weighted sum of $M_{1}$ and $M_{2}$. 

During the training phase, we increase the weight of the first branch to alleviate overfitting, and during the testing phase, we increase the weight of the second branch to improve prediction results. The first branch helps prevent the model from overfitting on the source domain because it has a simple structure and can learn stable and generalizable features during the training phase. The second branch provides sophisticated domain adaptation capabilities but due is more prone to overfitting when working alone.

\subsection{Adaptive Refine Self-Matching}
\begin{figure}[t]
  \centering
  \includegraphics[width=1.0\linewidth]{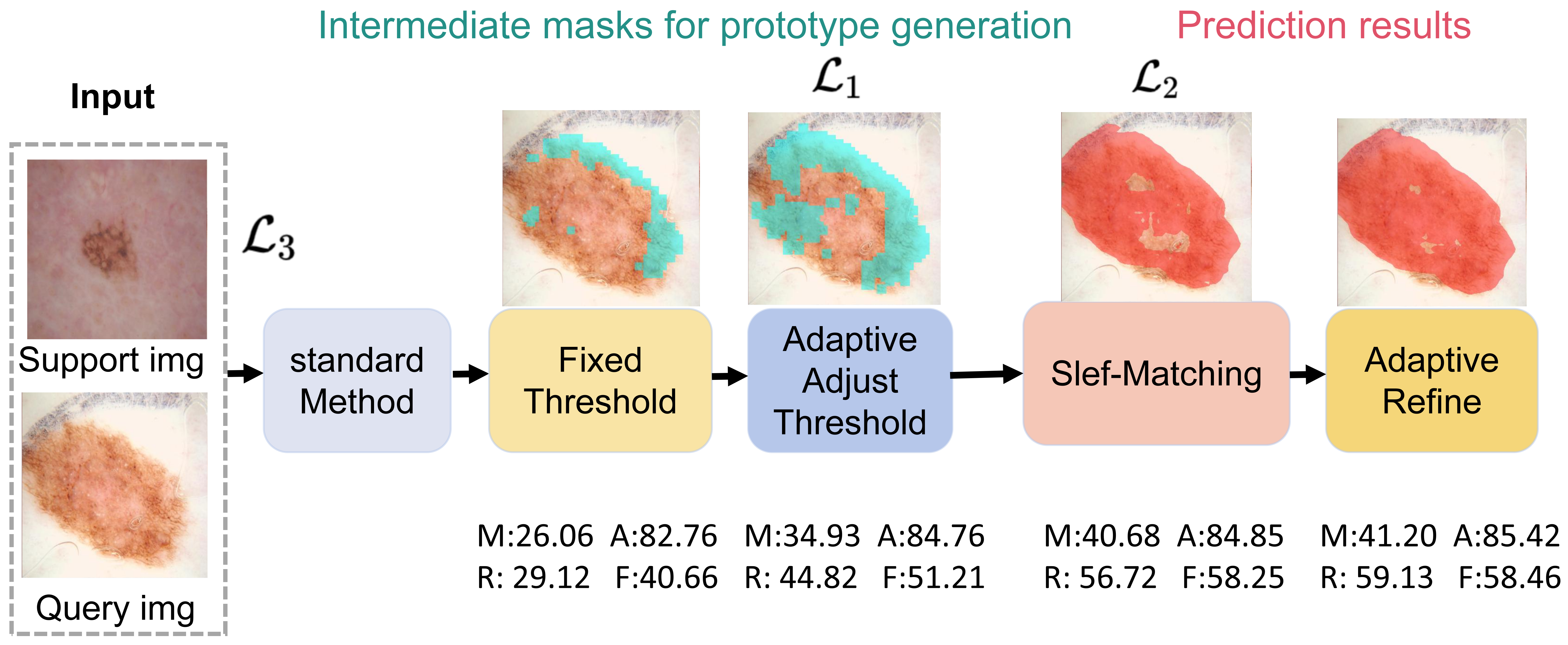}
  \caption{
   Visualization of our ARSM strategy.  We show the intermediate masks to filter query prototypes for self-matching, and the mIoU, accuracy, recall and F-score results of ARSM tested on the ISIC dataset.}
  \label{fig:image2}
\end{figure}

As Fig.~\ref{fig:image2} shows, the core idea of the Adaptive Refine Self-Matching (ARSM) method is to use the features that have been matched in the query image to generate a prototype and then to continue matching the query features  adaptively based on the target domain. This method is based on the intuition that feature similarity between various parts of an object is much greater than  similarity between features of different objects. In the original self-matching method \cite{yang2021mining}, based on the similarity map of the query features, we use a threshold to filter it to generate a prototype of the query feature. However, we notice that to ensure the effectiveness of self-matching,  it is crucial to adjust the threshold to include as few background features as possible while more comprehensively including the rich information of the foreground. Since the foreground and background of images in different target domains vary greatly, a single threshold does not perform well on data from different domains. Our method  combines the support set of the current target domain episode with the foreground and background similarity of the query set and the proportion of indistinguishable pixels in the query image to adaptively adjust the threshold. This helps solve the problem of reduced matching ability of prototype matching methods in cross-domain tasks.

We use the same method as the standard FSS segmentation task, combining $M_{s}$ and $F_{s}$ along with masked average pooling \cite{zhang2020sg} to obtain the prototype of the support images $P_s = \{ P_{s,f}, P_{s,b} \}$.
Cosine similarity is used to calculate the similarity $FB_{s}$ between the support foreground $P_{s,f}$ and the background $P_{s,b}$ prototype.
We then compute the matching confidence map prediction $M_{pre}=softmax(cosine(p_{s}, F_{q}))$, and  sum of confidence of each pixel on the foreground and background which should be 1.
We  threshold $M_{pre}$ to obtain a binary mask $\widetilde M_{pre}=M_{pre} I_{\left \{pixel>\tau    \right \} }$, with initial threshold $\tau =\{\tau_{f}=0.8, \tau_{b}=0.6\}$ to filter out the foreground and background of query features with higher confidence. Next, we combine $\widetilde M_{pre}$ and $F_{p}$ and use masked average pooling to obtain the initial query prototype $P_{q}=\{P_{q,f}, P_{q,b} \}$.
We also use cosine similarity to calculate the similarity between the query foreground and background prototypes to obtain $FB_q$.

However, the foreground and background confidence of each pixel in $M_{pre}$ can only indicate that each pixel is more like the foreground or more like the background; it cannot reflect relative  similarities between foreground or background. 
For some pixels, because their foreground and background cosine similarity values are very close, they do not exceed the threshold on the foreground or background and hence neither filtered into the foreground mask nor the background mask. We can also use thresholding to filter out  pixels that have not been classified as foreground or background in $\widetilde{M}_{pre}$; conversely we can get these pixels from $\widetilde{M}_{Lost} = M_{pre}I_{\left \{pixel<\tau    \right \} } $.
When we use thresholds to filter out foreground and background features, the filtering condition is the confidence of the background for each pixel. However, there are some pixels in $\widetilde{M}_{Lost}$ whose foreground and background similarity values are close and higher than the average similarity of the prototypes $P_{q}$, causing the foreground and background confidence of these pixels to be lower than the threshold. We can consider pixels that meet these conditions as the part of the query image that looks like both the foreground and background. 

In order to find these pixels, we need to calculate the average cosine similarity of  $P_{q}$ in foreground and background $ave = \{ave_{f}, ave_{b} \}$.
Then multiply $\widetilde{M}_{Lost}$ and $P_{q}$ to get the prototype of unclassified pixels and use $ave$ to filter pixels that are similar to both foreground and the background.
\begin{equation}
\label{eq: sim mask}
M_{similar} = P_{q}\widetilde{M}_{Lost} I_{\left \{Similarity > ave    \right \} } 
\end{equation}
After that, we can adaptively adjust the threshold by $\tau_{new} = \tau_{initial} + \delta$, $\tau_{initial}$ is 0.7, $\tau_{new}$ is the adjusted threshold, $\delta$ can be calculated according to the following equation:
\begin{equation}
\label{eq: adaptively adjust new}
\delta =  \Delta_{max} \ \left( \lambda \frac{{ FB_{q}- FB_{s}}}{{FB_{q}}} + (1 - \lambda )\frac{{Sim_{num}}}{{Union_{num}}} \right)
\end{equation}
$ \Delta_{max}$, the maximum value for adjustment is 0.2*$\kappa$, $Sim_{num}$ representing the number of pixels contained in $M_{similar}$. $Union_{num}$ is the number of pixels contained in the union of the foreground and background masks of the prototype $p_{q}$. $\kappa$.  $\lambda$ is a trainable parameter and its value is between 0 and 1. We will train this parameter at test time to make these parameters more consistent with the target domain. We can use $\tau_{new}$ to filter $M_{pre}$ to obtain a binary mask $\widetilde M_{new}=M_{pre} I_{\left \{pixel>\tau_{new}    \right \} }$ and use masked average pooling on $F_{p}$ obtain the new query prototype $P_{q}^{*}$.

In the first self-matching stage, the support prototype of the first stage SM is $P_{s}^{sm1} = \alpha P_{s} + \beta P_{q}^{*}$. $\alpha$ and $\beta$ are both 0.5 in the first stage SM of ARSM. We calculate the cosine similarity between the support prototype $P_{s}^{sm1}$ and the query feature $F_{q}$ to obtain the prediction $Out1$.
\begin{equation}
\label{eq: ssp1 out}
Out1 = softmax(cosine(P_{s}^{sm1}, F_{q}))
\end{equation}
ARSM continuously refines result $Out1$ by repeating the self-matching process. In each refining stage, we increase $\tau_{new,fg}$ by 0.05 and decrease $\tau_{new, bg}$ by 0.05 to more strictly screen the features of the foreground for self-matching.
After each stage of refinement we calculate the similarity between the foreground and background comparing it with the previous stage. If the similarity does not decrease, we stop the refinement and take the result of the previous stage. After refinement, we obtain prediction $M_{2}$.

\subsection{Channel Statistics Disruption}
Our CSD method is designed to densely and randomly perturb low-level features in the backbone to shift channel statistics of the source domain features toward the target domain. CSD can synthesize diverse regional styles by randomly changing the mean and standard deviation of feature channel statistics within each bottleneck of shallow layers, enhancing the generalization ability of the model.

Adjusting the mean and standard deviation of the input feature channel statistics can effectively change the style of the input image. AdaIN \cite{huang2017arbitrary} uses an adaptive affine transformation to adapt input to any given style.
The input parameters $x$ and $y$ of AdaIN are related to content and style respectively.$\left \{ \sigma (x), \mu (x)  \right \} \in R^{B\times C} $ and $\left \{ \sigma (y), \mu (y)  \right \} \in R^{B\times C} $ are feature channel statistics of the input content images and style images thereby generalizing the style of the extracted image features to unknown domains.
\begin{equation}
\label{eq: AdaIN}
  AdaIN(x,y)= \sigma (y)(\frac{x-\mu (x)}{\sigma (x)} )+\mu (y)
\end{equation}
In the case where the style of the target domain is not determined, we cannot get specific values for variables $\sigma (y)$ and $\mu (y)$ related to image style. In order to make the model more adaptable to random template domain styles, inspired by NP \cite{fan2022towards}, we replace $\sigma (y)$ and $\mu (y)$ with randomly scaled $\sigma (x)$ and $\mu (x)$ in our Channel Statistics Disruption (CSD) strategy.
\begin{equation}
\label{eq: CSD}
  CSD(x)= \lambda \sigma (x)(\frac{x-\mu (x)}{\sigma (x)} )+\delta \mu (x)
\end{equation}
where the coefficients $\lambda$ and $\delta$ can be regarded as random noise generated by The Normal distribution.

We incorporate CSD within each bottleneck of shallow layers in ResNet50 and set the noise ratio to 0.5 to enhance its ability to learn stable and robust features, improving generalization and reducing overfitting and sensitivity to data variability. If we add noise to deeper layers it may destroy the extraction of high-level semantic information.

\subsection{Test-time Adaptation}
We add a distribution alignment module (DAM) composed of efficient channel attention (ECA) \cite{wang2020eca} and 1×1 convolutional layer behind the backbone of the $M_{2}$ branch, and combine it with the learnable parameters in ARSM when implementing TTA. By fine-tuning the DAM module during the test phase, the DAM module focuses on those feature channels that are most critical to the segmentation task, ensuring that the model can adapt to the feature distribution of the target domain. We also train the two learnable parameters $\kappa$ and $\lambda$ in Eq. \ref{eq: adaptively adjust new} of ARSM section to dynamically adjust the threshold $\tau$ in ARSM based on each episode in different target domains. With 5-shot training, inspired by the idea of Prototype Alignment Regularization \cite{wang2019panet}, we select different query samples in each episode and use the remaining samples as the support set to dynamically adjust the prototype alignment target so that the information can effectively flow between the support and query sets to enhance the adaptability of the model.

In the test phase, $D_{test}$ is composed of episodes. 
Each episode consists of support set $S = \{(I_{i}^{s}, M_{i}^{s})\}_{i=1}^{N\times K}$ and a query set $Q = \{(I_{i}^{q}, M_{i}^{q})\}_{i=1}^Q$. At test time, we adapt the model with each episode, by updating the DAM module while freezing the other layers. We load the initial model every time we train each episode so that the training gradient does not accumulate, and we test the adapted model immediately after each episode training is completed. Our entire TTA strategy can be implemented as Algorithm \ref{alg:algorithm1}.

For the 1-shot task, we use a combination of colorjitter, blur, grayscale, and Cutout to perform data augmentation \cite{yang2022st++} on a support image to obtain a pseudo sample. We use this pseudo sample as a support sample and use the original support sample as a query sample to form an episode to fine-tune the DAM module and learnable parameters.

For the 5-shot task, we take turns using each sample in the support set $S$ as a query sample in each episode training and the other four samples as support samples to complete fine-tuning of the DAM  and learnable parameters. This strengthens the flow of information between samples in the support set and the mutual alignment of such prototypes helps to improve the generalization ability of the model. 

\begin{algorithm}
\footnotesize
\caption{Model Test-Time Adaptation Strategy}
\label{alg:algorithm1}
\begin{algorithmic}[1]
\For{each episode in $D_{\text{test}}$}
    \State Load initial model
    \State Freeze all layers except the DAM module
    
    \State Extract support set $S = \{(I_s^i, M_s^i)\}_{i=1}^{N \times K}$ and query set $Q = \{(I_q^i, M_q^i)\}_{i=1}^{Q}$
    
    \If{the task is 1-shot}
        \State Augment support image $I_s^1$ to obtain a pseudo sample $I_s^{\text{pseudo}}$
        \State Set $S_{1} = \{(I_s^{\text{pseudo}}, M_s^1)\}$
        \State Set $Q_{1} = \{(I_s^1, M_s^1)\}$
        
        \State Fine-tune DAM module and the learnable parameters in ARSM using $S_{1}$ and $Q_{1}$ by using $\mathcal{L}_{TTA}$
        \State Test the fine-tuned model with the query set $Q_{1}$
        
    \ElsIf{the task is 5-shot}
        \For{$i$ from 1 to 5}
            \State Select $I_s^i, M_s^i$ as a query sample and the remaining four as support samples
            \State Fine-tune DAM module and the learnable parameters in ARSM using the modified support and query sets by using $\mathcal{L}_{TTA}$
        \EndFor
        \State Test the fine-tuned model with the original query set $Q$
    \EndIf
\EndFor
\end{algorithmic}
\end{algorithm}

\subsection{Loss Function}
In the training phase of our DARNet, we calculate binary cross entropy (BCE) loss for the prediction results before self-matching $\mathcal{L}_{1} = BCE(Softmax(\widetilde M_{pre}), M_{gt})$, and the prediction results of self-matching $\mathcal{L}_{2} = BCE(Softmax(Out1), M_{gt})$. We also use the extracted support image features and prototypes to calculate binary cross entropy loss $\mathcal{L}_{3}$ for support images, to ensure the accuracy of the support prototype generated during the matching process. Finally, we train the model using the weighted loss function: $\mathcal{L}_{train} = 1.0* \mathcal{L}_{1} + 1.0*\mathcal{L}_{2} + 0.3* \mathcal{L}_{3}$.

In the test-time adaptation phase we calculate the BCE loss $\mathcal{L}_{BCE} = BCE (\mathbf{p}, \mathbf{m})$, where $\mathbf{p}$ is the predicted probabilities, $\mathbf{m}$ is the ground truth for the support image. We also generate an attention map based on $M_{similar}$ of the indistinguishable areas in the ARSM method and design an attention loss function based on the attention map:
\begin{equation}
\mathcal{L}_{\text{att}}(\mathbf{p}, \mathbf{m}, \mathbf{A}) = 
\frac{1}{N} \sum_{i=1}^{N} \frac{\sum_{h,w} \left( 1 + \alpha \mathbf{A}_{i} \right) 
\cdot \text{BCE}(\mathbf{p}_{i}, \mathbf{m}_{i})}{\sum_{h,w} \left( 1 + \alpha \mathbf{A}_{i} \right)}
\end{equation}
$\mathbf{A}$ is attention map tensor. TTA on model by  weighted loss function gives: $\mathcal{L}_{TTA} = 0.7*\mathcal{L}_{BCE} + 0.3*\mathcal{L}_{\text{att}} $.

\begin{figure}[t]
  \centering
  \includegraphics[width=1.0\linewidth]{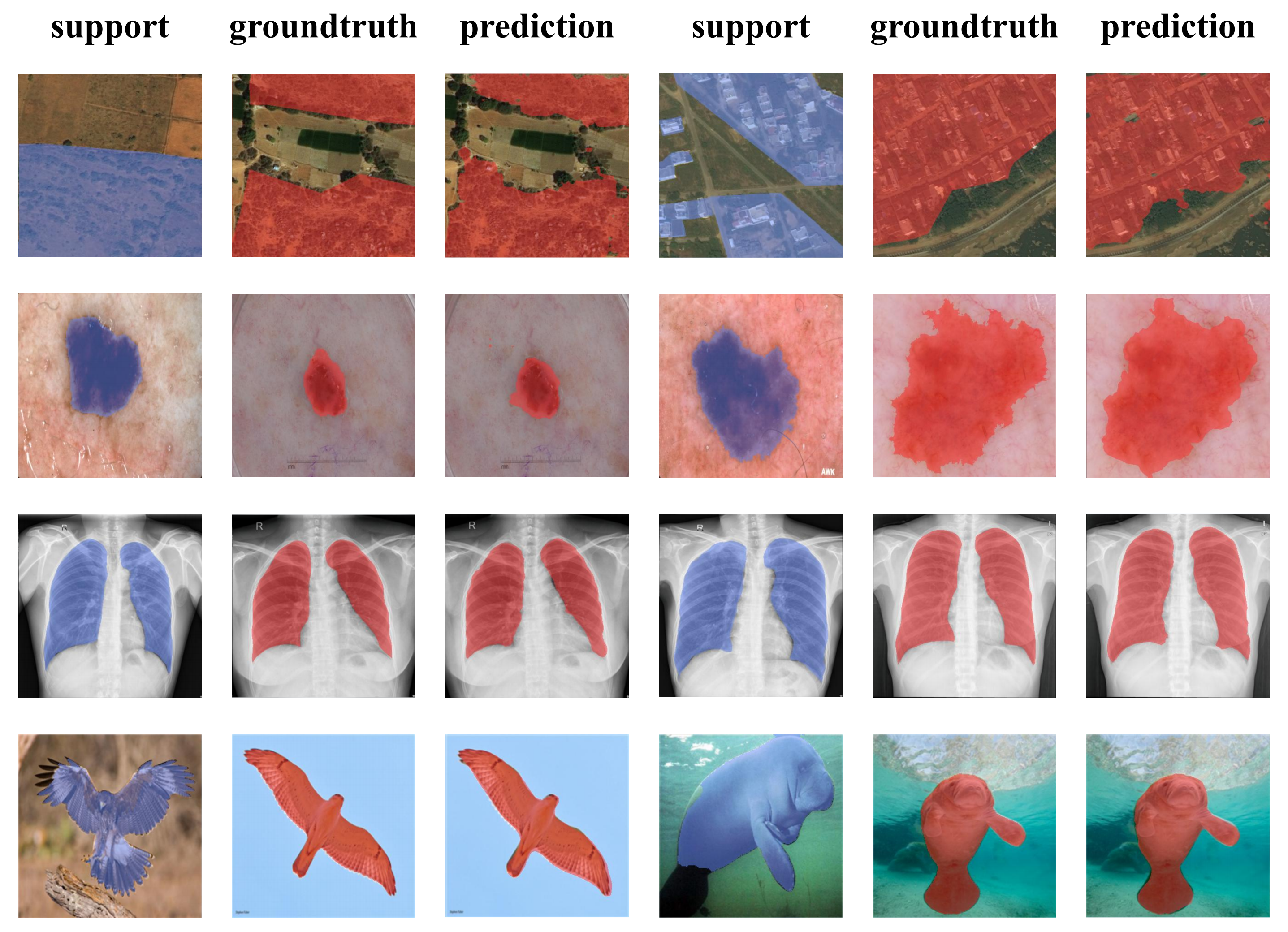}
  \caption{
From top to bottom, each row of 1-shot prediction results of our method comes from the datasets Deepglobe, ISIC, X-Ray, and FSS-1000. We overlay the support image labels in blue and the prediction results and query image labels in red. }
  \label{fig:all}
\end{figure}

\section{Experiment}

\subsection{Evaluation Setup}

In alignment with prior methods, we trained our models using the datasets PASCAL VOC 2012 \cite{everingham2010pascal} and SBD \cite{hariharan2011semantic} before testing on ISIC \cite{codella2019skin, tschandl2018ham10000}, Chest X-ray \cite{candemir2013lung, jaeger2013automatic}, Deepglobe \cite{demir2018deepglobe} and FSS-1000 \cite{li2020fss}.

FSS-1000 \cite{li2020fss} encompasses a total of 1,000 distinct categories of natural images, each provided with 10 exemplars. Evaluation of our models is carried out on the designated test set, which is comprised of 240 unique object classes and a total of 2,400 images.

Deepglobe \cite{demir2018deepglobe}, consists of satellite imagery, densely annotated at a pixel level across seven different categories which include a category termed `unknown'. Each image within this dataset is of a consistent spatial resolution, specifically 2448 × 2448 pixels. In an effort to augment the quantity of images available for testing, as well as to reduce the individual image sizes, we partition each image into six separate sections.  Following the exclusion of images labeled with a single class or the `unknown' category, we obtained a total of 5,666 images for evaluation, with image's dimensions reduced to 408 × 408 pixels.

ISIC2018 \cite{codella2019skin}, is comprised of three categories of lesion imagery used for skin cancer screening, which includes a total of 2,596 images. These images initially have a resolution of approximately 1022 × 767 pixels, but we reduce the image size to 512 × 512 pixels for analysis.

Chest X-ray \cite{candemir2013lung}, comprises 566 X-ray images of one category utilized for Tuberculosis analysis, originally with a resolution of 4020 × 4892 pixels. We resize them to a more manageable resolution of 1024 × 1024 pixels.

We calculate the mean Intersection-over-Union (mean-IoU) by averaging the results across five independent runs, each initiated with a random seed. For all datasets except FSS-1000, each run encompasses 1,200 tasks. For the FSS-1000 dataset, we use 2,400 tasks for each run, adhering to the configuration described in previous studies \cite{lei2022cross}. For each task, we tried 1-way 1-shot and 1-way 5-shot.

\subsection{Implementation Details}

As in previous methods \cite{lei2022cross}, we use ResNet50 \cite{he2016deep} as our backbone for feature extraction.
Unlike previous methods  \cite{lei2022cross}, we do not use conv5\_x of ResNet50 but rather use the feature map generated by conv4\_X, and the channel dimension of this layer is 1024.  Following the previous method, we set the size of the support images and query images to 400 × 400 in the training and TTA phases. This model was implemented using pytorch, and we use Adam optimizer to train and fine-tune the model with a learning rate of 1e-3. For the corresponding TTA task on each target domain dataset, we perform 10 iterations.

\begin{table*}[t]
\footnotesize
\centering

\begin{tabular}{|l|cc|cc|cc|cc|cc|cc|}
\hline
\multirow{2}{*}{Methods} & \multicolumn{2}{c|}{Deepglobe}& \multicolumn{2}{c|}{ISIC} & \multicolumn{2}{c|}{Chest X-ray} & \multicolumn{2}{c|}{FSS-1000} & \multicolumn{2}{c|}{Average}\\
\cline{2-11}

& 1-shot & 5-shot & 1-shot & 5-shot & 1-shot & 5-shot & 1-shot & 5-shot & 1-shot & 5-shot\\

\hline
PGNet \cite{siam2019amp} & 10.73 & 12.36& 21.86 & 21.25 & 33.95 & 27.96 & 62.42 & 62.74 & 32.24 & 31.08 \\
PANet \cite{zhang2019pyramid}  & 36.55 &\underline{45.43} &25.29 &33.99 &57.75 &69.31 &69.15 &71.68 &47.19 &55.10\\
CaNet \cite{zhang2019canet} &22.32 &23.07 &25.16 &28.22 &28.35 &28.62 &70.67 &72.03 &36.63 &37.99 \\
RPMMs \cite{yang2020prototype} &12.99 &13.47 &18.02 &20.04 &30.11 &30.82 &65.12 &67.06 &31.56 &32.85\\

PFENet \cite{tian2020prior} &16.88 &18.01 &23.50 &23.83 &27.22 &27.57 &70.87 &70.52 &34.62 &34.98\\

RePRI \cite{boudiaf2021few} &25.03 &27.41 &23.27 &26.23 &65.08 &65.48 &70.96 &74.23 &46.09 &48.34\\

HSNet \cite{min2021hypercorrelation} &29.65 &35.08 &31.20 &35.10 &51.88 &54.36 &\underline{77.53} &80.99 &47.57 &51.38\\

SSPNet \cite{fan2022self} &35.09 &36.39 &37.68 &38.12 &\underline{72.33} &\underline{73.80} &75.50 &76.81 &55.15 &58.78\\

PATNet \cite{lei2022cross} &\underline{37.89} &42.97 &\underline{41.16} &\underline{53.58} &66.61 &70.20 &\textbf{78.59} &81.23 &\underline{56.06} &\underline{61.99} \\
\hline
DARNet(Ours) &\textbf{44.61} &\textbf{54.05} &\textbf{47.81} &\textbf{60.52} &\textbf{81.22} &\textbf{89.73} &76.41&\textbf{83.24} &\textbf{62.51} &\textbf{71.89} \\

\hline
\end{tabular}
\caption{MIoU results of few-shot segmentation methods on 1-way 1-shot and 5-shot results on CD-FSS tasks. All methods are trained on PASCAL VOC and tested with CD-FSS.}
\label{table1}
\end{table*}

\begin{figure}[t]
  \centering
  \includegraphics[width=1.0\linewidth]{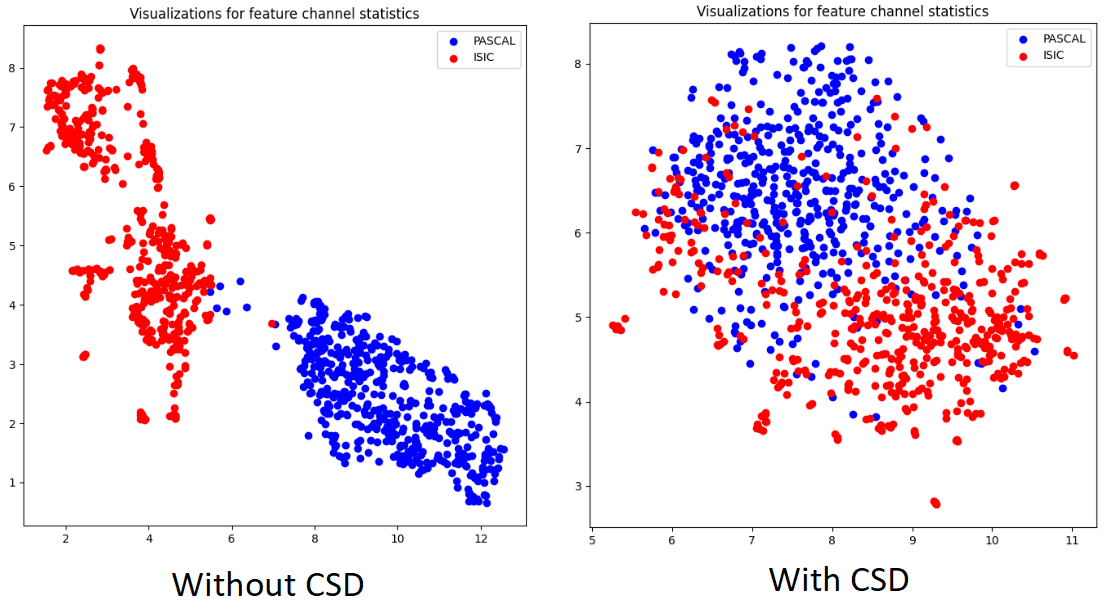}
  \caption{
   We use blue and red scatter points to represent features of the source domain dataset PASCAL and the target domain dataset ISIC after backbone processing with and without CSD strategy.}
  \label{fig:csfss}
\end{figure}

\subsection{Comparison with State-of-the-art}
As shown in Table \ref{table1}, Our method outperforms all few-shot segmentation methods except the 1-shot task on the FSS-1000 dataset which demonstrates the advantages of our method on some datasets with large domain gaps. The average mean-IoU of our 1-shot and 5-shot on the four datasets are 6.45\% and 9.90\% higher than the second-place method PATNet respectively. PATNet uses the Pyramid Anchor-based Transformation Module (PATM) to transform features learned from the source domain data set into domain-agnostic features. Similarly, our CSD strategy generalizes the learned features and fits them to the feature distribution of the unknown target domain by densely perturbing low-level features, offering speed and fewer parameters. In addition, PATNet fine-tuned the weight anchor layer of PATM during testing by using the similarity between the prototype of the support set and the predicted prototype of the query set as the loss function to solve the cross-domain gap problem. When fine-tuning, they assume that the foreground of the query set and the support set are similar while ignoring the intra-class gap problem in the same data set, which also leads to poor inference results of PATNet. Our method solves the intra-class gap and cross-domain gap problems by combining ARSM and TTA. However, since the CSD strategy will destroy the feature distribution of the source domain to a certain extent, our method does not work the best on the FSS-1000 data set with a small domain gap.
We present some results of our method in Fig. \ref{fig:all}.

\begin{table}[ht]
\footnotesize
\centering

\begin{tabular}{|c|c|c|c|c|c|c|}
\hline
baseline & SM & CSD & ARSM & TTA & 1-shot & 5-shot \\
\hline
$\checkmark$ & & & & & 52.54 & 55.54 \\
\hline
$\checkmark$ & $\checkmark$ & & & & 55.15 & 58.78 \\
\hline
$\checkmark$ & $\checkmark$ & $\checkmark$ & & & 57.30 & 62.62 \\
\hline
$\checkmark$ & $\checkmark$ & $\checkmark$ &  $\checkmark$ & & 58.50 & 64.03 \\
\hline
$\checkmark$ & $\checkmark$ & $\checkmark$ & $\checkmark$ & $\checkmark$ &  62.51 & 71.89 \\
\hline
\end{tabular}
\caption{Ablation study with average 1-shot and 5-shot performance. Our baseline model is a mainstream method MLC \cite{yang2021mining}.}
\label{table:ablation_study}
\end{table}

\begin{figure}[t]
  \centering
  \includegraphics[width=1.0\linewidth]{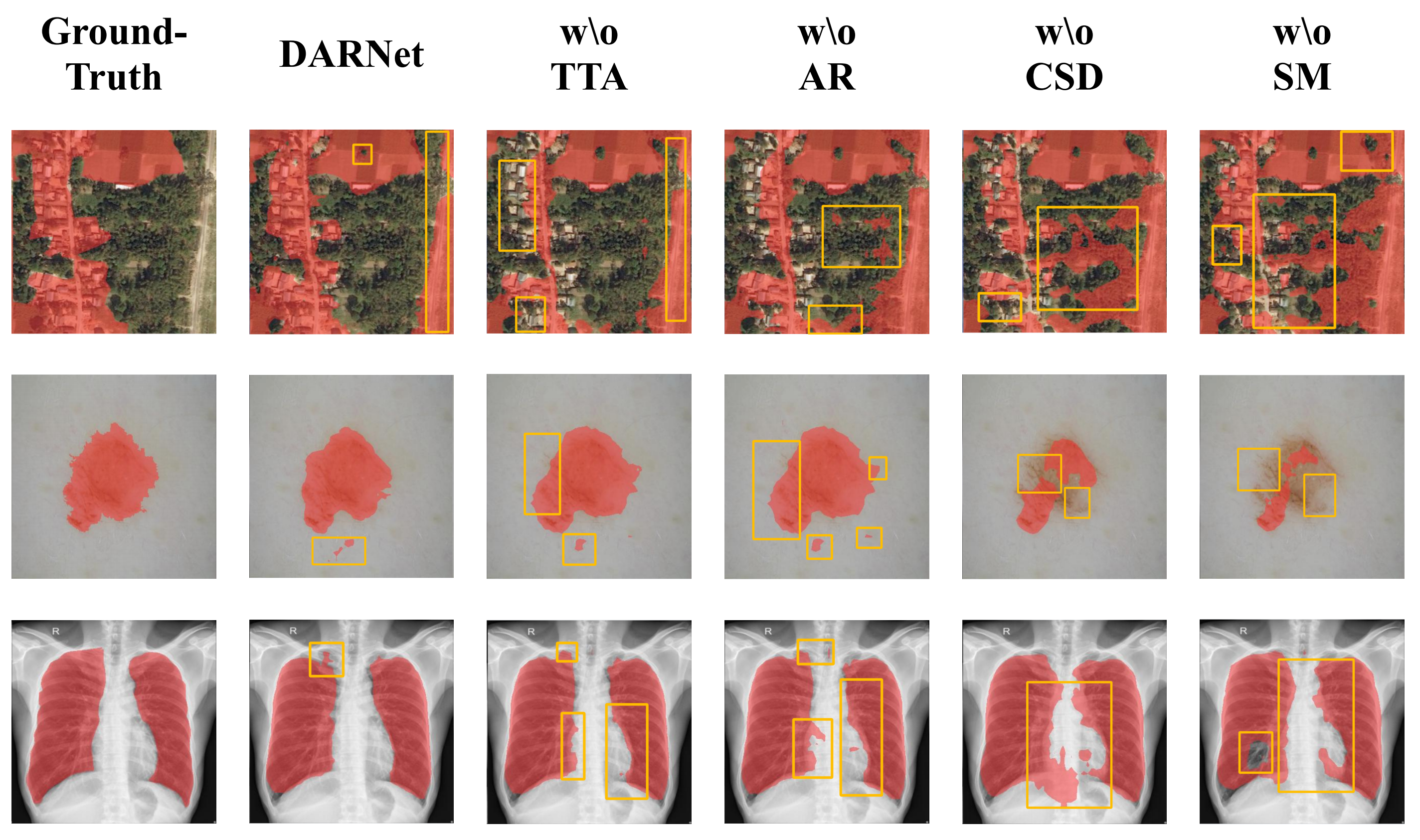}
  \caption{1-shot results of our method on several tasks with large domain gap, showing ablation experimental results for all modules in our method, with yellow boxes marking areas with obvious changes.}
  \label{fig:ab}
\end{figure}

\subsection{Ablation Study}
We conducted a series of ablation experiments to study the performance of the Self-Matching method and the impact of CSD, ARSM, and TTA strategies. Fig. \ref{fig:ab} shows the impact of all modules on the prediction results.

\textbf{Performance of Self-Matching (SM)}.
First, we studied the performance of the Self-Matching few-sample segmentation method on the CD-FSS task. From Table \ref{table:ablation_study}, we can find that the performance of SM on the CD-FSS data set is almost close to that of PATNet.

\begin{table}[ht]
\footnotesize
\centering

\begin{tabular}{|c|c|c|c|c|c|c|}
\hline
L$_{3}$ & L$_{2}$ & L$_{1}$ & L$_{1 \uparrow n} $  & B$_{1}$  & 1-shot & 5-shot\\
\hline
$ \times $ & $ \times$ & $ \times$ & $ \times$ & $ \times$& 54.15 & 58.78 \\
\hline
$\checkmark$ & $ \times $& $ \times $ & $ \times$ &$ \times $&55.08  &60.18  \\
\hline
$ \times $ & $\checkmark$ & $ \times $ & $ \times$ & $ \times $&55.19  &59.77  \\
\hline
$ \times $ & $ \times $ & $\checkmark$ & $ \times$ &$ \times $ & 55.81 & 60.40 \\
\hline
$ \times $ & $ \times $ & $ \times$ & $\checkmark$ &$ \times $ & 54.93 & 59.46 \\
\hline
$ \times $ & $ \times $ & $ \times $ & $ \times$ &  $\checkmark$ & 57.30 & 62.62 \\

\hline
\end{tabular}
\caption{Ablation study on the choice of CSD position and noise ratio.}
\label{table:ablation_study_CSD}
\end{table}

\textbf{Effect of channel statistics disruption (CSD)}.
 In Table \ref{table:ablation_study}, we can see that adding CSD strategy improves 1-shot and 5-shot performance by 3.15\% and 2.84\% respectively. In addition, we also conducted quantitative experiments in Table \ref{table:ablation_study_CSD} on where the CSD strategy implements feature perturbations in the backbone. We define layer1, layer2, and layer3 of Resnet50 respectively, and show the average performance of using the feature perturbation method behind these three layers and in each block of the low layer. We also compared the results of increasing the noise ratio from 0.5 to 0.75 behind Lyaer1, and found that blindly increasing the noise ratio will cause the index to drop. From Table \ref{table:ablation_study_CSD}, it is evident that applying feature perturbation to the block of the lower layer yields optimal results. Fig. \ref{fig:csfss} shows CSD can effectively reduce the feature distribution gap between the source domain and the target domain.

\begin{table}[ht]

\centering
\footnotesize

\begin{tabular}{c|ccccc}
\hline
Strategy & Deepglobe & ISIC & X-Ray & FSS-1000 & Average  \\
\hline
Initial(0.8)  & 39.35  & 39.84 & 74.43  & 75.41  & 57.30 \\
$\uparrow$ 1.0 &38.79 &38.81 &74.90 & 75.62 & 57.03 \\
$\uparrow$ 0.5 &39.16 &39.43 &75.43 &75.78  & 57.45 \\
$\downarrow$ 0.5 &40.22 &40.08 &72.17 &72.65 & 56.28  \\
$\downarrow$ 1.0 &39.89 &40.35 &71.58 &72.26  & 56.02 \\
Adaptive & 41.45  & 41.20 & 75.48  & 75.87 & 58.50  \\
\hline
\end{tabular}
\caption{Ablation study on adaptive and manually adjusted thresholds with 1-shot performance on all CD-FSS tasks. In this study we use the model without TTA.}    
\label{table: AR}
\end{table}
\textbf{Effect of adaptive refine Self-Matching (ARSM)}.
From Table \ref{table:ablation_study}, we can find that by using the ARSM strategy, the performance of 1-shot and 5-shot is improved by 1.2\% and 2.41\% respectively. In Table \ref{table: AR}, we also compare the average performance of our ARSM strategy, no adjustment with the threshold of 0.8, and the method of manually adjusting the threshold on all test sets. We can see that the adjusting threshold specifically for each episode is better than simply manual adjustment. This reflects that our ARSM strategy can take advantage of the difference in foreground and background similarity between support and query in each episode to better dynamically refine prediction results.

\begin{table}[ht]
\footnotesize
\centering

\begin{tabular}{c|cc}
\hline
Structure & 1-shot & 5-shot \\
\hline
1*1conv &60.02  &71.15  \\
3*3conv &59.34  &71.02  \\
2$\times$1*1conv & 58.9 & 70.57 \\
2 $\times$3*3conv & 58.63 &70.01  \\
ASPP \cite{chen2017deeplab}+1*1conv & 61.83 & 71.71 \\
PPM \cite{zhao2017pyramid}+1*1conv & 60.91 & 70.59 \\
DAM & 62.51 & 71.89 \\

\hline
\end{tabular}
\caption{Ablation study on different structures for TTA.}       
\label{table:TTA}
\end{table}

\textbf{Effect of Test-time-adaptation (TTA)}.
After adding the TTA strategy, it can be found from Table \ref{table:ablation_study} that the performance of 1-shot and 5-shot has improved by 4.01\% and 7.86\% respectively. This shows that our TTA strategy allows the model to better adapt to the data in the target domain and ultimately achieve the best performance. In Table \ref{table:TTA}, we find that simply stacking convolutional layers degrades performance. The performance of ASPP and PPM using multi-scale methods is not as good as that of the ECA module using channel attention methods. This shows that adjusting the weight of feature channels is better than using multi-scale methods focusing on contextual relationships.

\section{Conclusion}
In this paper, we propose a Dynamic Adaptive Refinement Network (DARNet), which simultaneously addresses two major challenges in cross-domain few-shot segmentation (CD-FSS) tasks: inter-domain differences and intra-class inconsistency. Through channel statistical interference and adaptive strategies, DARNet not only improves the generalization ability from the source domain to the target domain, but also achieves more accurate predictions in different target domains. Evaluated on four CD-FSS tasks using four different target domain datasets, our experimental results show that DARNet achieves consistently better performance than existing state-of-the-art methods, confirming the advantages of our method for both few-shot segmentation and generalizability for different domains and tasks.

{
    \small
    \bibliographystyle{ieeenat_fullname}

}


\end{document}